%% file: arxiv.tex
\documentclass{article} 
\usepackage{iclr2027_conference,times}

\input{math_commands.tex}

\PassOptionsToPackage{hyphens}{url}
\usepackage{hyperref}
\usepackage{url}
\usepackage{graphicx}
\usepackage{float}
\usepackage{amsmath,amssymb}
\usepackage{booktabs}
\usepackage{xcolor}
\usepackage{soul} 

\graphicspath{{figures/}}
\newcommand{\Mh}{\mathcal{M}_h}
\newcommand{\My}{\mathcal{M}_y}
\newcommand{\dmodel}{d_{\mathrm{model}}}

\title{Spokes, not a Ring: Revisiting the Weekday Manifold in Language Models}

\author{Diksha Gupta \\
Pivotal Research \\
\And
Rieke Fruengel\\
Pivotal Research \\
\And
Stefan Heimersheim \\
}

\iclrfinalcopy 
\begin{document}

\maketitle

\begin{abstract}
Recent work has hypothesised that large language models may represent some concepts as multi-dimensional manifolds instead of single directions. Using these manifolds to interpret or steer representations often assumes that moving along them corresponds to a continuous change in the represented concept. In this work, we test this continuous manifold assumption on the prototypical weekday manifold. Despite its striking circular geometry, we find that the weekday representations behave more like seven discrete states arranged in a circle than a continuous ring. First, we show that local sensitivity has seven minima aligned with the weekdays, producing a step-like mapping between representation and behaviour under on-manifold steering. Second, we test whether any concepts are represented along the manifold outside the seven discrete states, and find neither time of day, uncertainty, nor natural text populate the intermediate space. Together, these results challenge the continuous interpretation of the weekday manifold and motivate testing whether movement along other representation manifolds correspond to meaningful changes in model representations and behaviour. 
\end{abstract}

\section{Introduction}
\label{sec:intro}

A growing body of work proposes that some concepts are represented by language models not as single directions in activation space but as low-dimensional geometric structures or manifolds. The canonical example is the days-of-the-week ring: activations for the seven days lie in a low-dimensional subspace and are arranged in calendar order around a circle \citep{engels2025notall,modell2025origins,wurgaft2026manifold}. Similar geometric structures have been reported for spatio-temporal \citep{gurnee2024space}, numerical \citep{kantamneni2025trigonometry}, belief-like  \citep{shai2024belief,sarfati2026shape}, and many others concepts \citep{tiblias2026manifold, bhalla2026manifolds}. The manifold view is appealing because geometry in activation space can reflect semantic relationships between the underlying concept values. For example, real-world relationships between places and times can be recovered from the activation space of language models \citep{gurnee2024space,tiblias2026manifold,modell2025origins}. 

In practice, however, these geometric structures are inferred from a finite set of observed representations. A continuous manifold is then fit through those points, so the regions between them are determined by interpolation. The interpolated regions may correspond to representations of intermediate concept values, such that semantic structure extends continuously across the manifold. Alternatively, they may simply connect discrete representational states, causing steering along the fitted manifold to pass through activation regions that are geometrically well defined but have no meaningful semantic interpretation. We refer to the assumption that points throughout the manifold correspond to meaningful values of the represented concept as the \emph{continuous manifold assumption}. 

\begin{figure}[!t]
\begin{center}
\includegraphics[width=\linewidth]{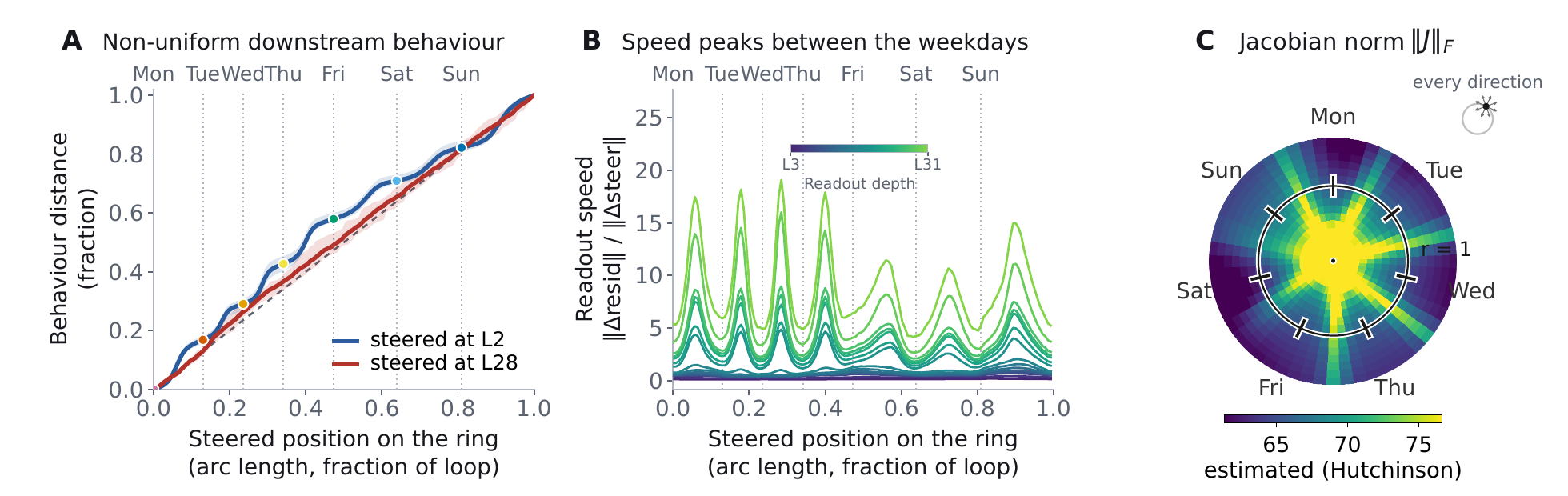}
\end{center}
\caption{\textbf{Perturbation sensitivity around the weekday ring has seven weekday-centred minima.} 
(A) Early-layer steering produces step-like changes in behaviour, with little change near weekdays and rapid change along the intervening arcs. We steer at constant angular speed around the fitted ring and measure cumulative distance in behaviour space. Steering at layer 2 (blue) produces a step-like response whereas steering at layer 28 (red) is smoother.
(B) The difference in downstream response speed between weekday positions and the intervening arcs becomes more pronounced over layers. We steer around the ring at layer 2 and measure the speed of the residual-stream response at layers 3--31.
(C) The sensitivity minima extend radially beyond the fitted ring, forming seven weekday-centred spokes separated by regions of higher sensitivity. At each point in the surrounding activation space, we measure how much a small change to the layer-2 activation changes the layer-28 residual stream, using the Frobenius norm of the corresponding Jacobian.}
\label{fig:sensitivity}
\end{figure}



The days-of-the-week ring provides a natural setting in which to examine this assumption. \citet{engels2025notall} find that earlier times shift the representation towards the previous day, while later times shift it towards the following day, suggesting that the weekday ring\footnote{For brevity we refer to the days-of-the-week ring as the \emph{weekday ring} throughout. ``Weekday'' here means any of the seven days, including Saturday and Sunday.} may represent time continuously between neighbouring days. \citet{wurgaft2026manifold} fit a smooth curve through the seven weekday representations at a late layer and observe smooth behavioural changes when steering along it. However, whether the interpolated regions of the weekday ring are genuinely part of the model's representation remains unresolved. Establishing representational continuity requires evidence that intermediate activations occur naturally (i.e. are in-distribution) and that intermediate values affect downstream behaviour in ways consistent with intermediate concept values. 

Therefore, here we explicitly test the continuous manifold assumption on the weekday ring and find evidence favouring seven discrete states over a continuous ring. We perform two complementary analyses. First, motivated by prior work showing that in-distribution activations can occupy regions of low perturbation sensitivity \citep{janiak2024stable,shinkle2025plateaus}, we examine whether weekday representations exhibit this structure and whether steering along the arcs produces gradual or step-like changes in downstream behaviour. Second, we investigate whether the arcs of the weekday ring encode meaningful intermediate concept values by examining activations elicited by time of day, uncertainty between weekdays, and natural text.

Our main contributions are:
\begin{itemize}
  \item We show that activations along the weekday ring don't map to behaviour continuously but in a step-function like manner, with only seven discrete regions along the ring showing signatures of in-distribution activations (low Jacobian norms).
  \item We show that the arcs between weekdays neither causally encode time of day nor represent uncertainty between weekdays.
  \item We show that activations populate 7 clusters (spokes) along the ring, but not the space in between (arcs). This clustering is independent of context: different natural language contexts preserve this discrete structure cause global shifts in activations, but no rotation along the ring.
\end{itemize}

These findings motivate explicit tests of continuity in other proposed representational manifolds. Similar departures from continuity could inform both how such geometry is interpreted in feature discovery and whether it provides a meaningful path for steering.

\section{Related Work}
\label{sec:related}

\paragraph{Geometric structure in representations}
Much of mechanistic interpretability has modelled concepts as directions in activation space, but a growing body of work has identified richer low-dimensional geometry, including circular representations in modular arithmetic \citep{liu2022grokking,nanda2023progress}, circular representations of weekdays and months \citep{engels2025notall,wurgaft2026manifold}, helical representations of numbers \citep{kantamneni2025trigonometry}, structured representations of space and time \citep{gurnee2024space}, and colour representations that reflect perceptual colour geometry \citep{abdou2021color}. Interventions that follow or manipulate this geometry can systematically change model behaviour \citep{engels2025notall,wurgaft2026manifold,sarfati2026shape}. However, similar geometric structure can arise from statistical regularities in the data \citep{karkada2026symmetries,prieto2026data} and cyclic concepts can be represented circularly even when the model does not use that circle to perform the computation \citep{feucht2026arithmetic}. Together, these results motivate distinguishing the observed geometry from the semantic and computational structure inferred from it.

\paragraph{Representation manifolds and continuity}
Recent work has formalised the idea of a representation manifold, relating geometry in activation space to structure in an underlying feature space \citep{modell2025origins}. Continuous manifold structure has been used in probing and intervention \citep{wurgaft2026manifold,modell2026probing}, and treated as an object of recovery in feature discovery \citep{bhalla2026manifolds}. Other work provides examples where the represented quantity is explicitly continuous or distributional, including belief manifolds \citep{shai2024belief,sarfati2026shape}. Our work instead asks when continuity of the fitted geometry is supported by continuity in the underlying representation.

\paragraph{Local stability in activation space}
Perturbation studies have shown that in-distribution activations can occupy locally stable regions in which small interventions have little downstream effect, separated by boundaries across which behaviour changes rapidly \citep{heimersheim2024plateaus,janiak2024stable,humayun2024grok}. We use this local stability structure as a diagnostic for whether the weekday ring contains distinct representational states.

\section{Methods}
\label{sec:methods}

We study Llama-3.1-8B, Mistral-7B-v0.1 and GPT-2 XL (Appendix~\ref{app:setup}). Results in the main body are from Llama-3.1-8B. We construct the weekday manifold following \citet{wurgaft2026manifold}. We first average residual-stream activations across templated prompts (Appendix~\ref{app:prompts}) for each weekday to obtain seven centroids. We then fit a periodic cubic spline through these centroids in weekday order to obtain the days-of-the-week ring (Appendix~\ref{app:manifold}). 

\paragraph{Sensitivity:}
Prior work on activation plateaus \citep{shinkle2025plateaus}, shows that in-distribution regions of activation space have a lower sensitivity to perturbations, while out-of-distribution activations have higher sensitivity. We use their sensitivity metric (specifically, Jacobian norm) to evaluate which parts of the fitted manifold appears in-distribution. We expect computation to be spread across the layers of the model, with early and middle layers representing the day of the week concept and late layers representing the output prediction of the model. Hence we measure the activation-behaviour mapping at an early (layer 2) as well as late (L28) layer.

We use 20 third-person templates $\times$ 7 days such as ``On \{day\}, she missed her train.'' (Appendix~\ref{app:prompts}). We steer at the weekday token at layer-2 and read at the sentence-final token at 87.5\% depth (layer-28 in Llama and Mistral, layer-42 in GPT-2 XL). This differs from \citet{wurgaft2026manifold}, who steer close to the readout (layer-28) using arithmetic prompts such as ``Q: What day is two days after Monday?\textbackslash nA:''. We use third-person templates rather than the arithmetic prompts of \citet{wurgaft2026manifold}, since the latter strongly constrain the model to output a weekday at the readout position, which may enforce a discrete representation. The seven weekday tokens account for 0.739 of the next-token probability at their answer slot, against 0.021 at the readout position in our third-person templates. As a robustness check, Appendix~\ref{app:arithmetic} repeats the sensitivity measurements on their prompts at both intervention sites.

We first steer in equal steps around the fitted ring at layer-2 and measure the resulting change in the output distribution and late-layer residual stream (Appendix~\ref{app:steering}, Appendix~\ref{app:readout}). We then measure local sensitivity using the Jacobian of the late-layer residual with respect to the steered activation, using its Frobenius norm as a direction-independent measure of sensitivity. We evaluate this over a polar grid spanning the ring and surrounding activation space (Appendix~\ref{app:sensitivity}). Additional analyses test sensitivity across perturbation directions and whether the output-space pattern can be explained by weekday-related logits (Appendix~\ref{app:confound}).

\paragraph{Time of day:}
We test whether positions between neighbouring weekdays encode time within a day. We prompt with ``It was \{time\} on \{day\}.'' for 22 clock hours and seven weekdays and read the activation at the weekday token (Appendix~\ref{app:timeofday}). PCA on these activations is used to compare the dominant direction capturing variance due to clock-time and weekdays. 

To measure the clock-time component within the weekday plane, we project each activation into the weekday plane and compute its angular displacement from the corresponding weekday centroid. We repeat this analysis with broader temporal modifiers and non-temporal controls.

We then test whether movement along the fitted ring causally affects time-of-day predictions. We use prompts that end immediately before an hour prediction, for example ``The train leaves on Wednesday at around'', and then compute model-predicted hour from the next-token distribution (Appendix~\ref{app:timeofday}). At each layer, we steer the weekday-token activation towards an adjacent day along the ring ($\alpha=0.5$ denotes the halfway point) and measure the median change in predicted hour. We compare this with steering along the off-ring component of the observed ``late--early'' displacement and a random direction orthogonal control, both matched to the along-ring intervention in $\ell_2$ norm.

\paragraph{Uncertainty:}
We next test whether positions between weekdays encode uncertainty between candidate days. We construct prompts expressing graded uncertainty between weekday pairs (one, two, or three days apart). Across several prompt families, we vary how strongly the prompt favours one day over the other and read the residual-stream activation at the final period (Appendix~\ref{app:uncertainty}), for example, ``\{Usually\} on \{Monday\}, \{sometimes\} on \{Tuesday\}, her flight leaves.'' We then examine how the representation changes as uncertainty varies. We consider the hypotheses (i) uncertainty between two dates moves along the weekday manifold (``ring''), and (ii) an uncertain representation corresponds to exactly one day's representation but with a lower magnitude (``spokes''). For day pairs that produce multiple activation modes, we also analyse the modes separately to distinguish genuine interpolation from an intermediate mean produced by averaging day-aligned clusters.

\paragraph{Natural-text occupancy:}
Finally, we test whether natural text places weekday representations in the interpolated regions of the ring. From FineWeb \citep{penedo2024fineweb}, we take excerpts of exactly 16 tokens ending in a weekday ($49{,}678$ excerpts from $574{,}217$ documents; Appendix~\ref{app:naturaltext}). Equal-length excerpts ending in another capitalised word serve as controls. We read activations at the weekday token where the circular geometry is most pronounced (Appendix~\ref{app:ringtrace}). We fit the weekday subspace on a random third of the excerpts ($16{,}879$) and analyse the remaining $32{,}799$. 

To examine context-dependent rotations around the ring, we construct six matched variants of every sequence by replacing the final weekday with each of the other weekdays. This gives seven sequences with identical preceding context. We compare the resulting seven-day configurations across contexts, and subtract their within-context mean to remove common translational shifts. We visualise the seven-day configurations before and after mean subtraction in the same plane, and measure their alignment with the weekday directions.

Code to reproduce experiments and figures in this paper is available at \url{https://github.com/dikshagup/spokes-not-a-ring.git}.

\section{Results}
\label{sec:results}

\begin{figure}[!t]
\begin{center}
\includegraphics[width=\linewidth]{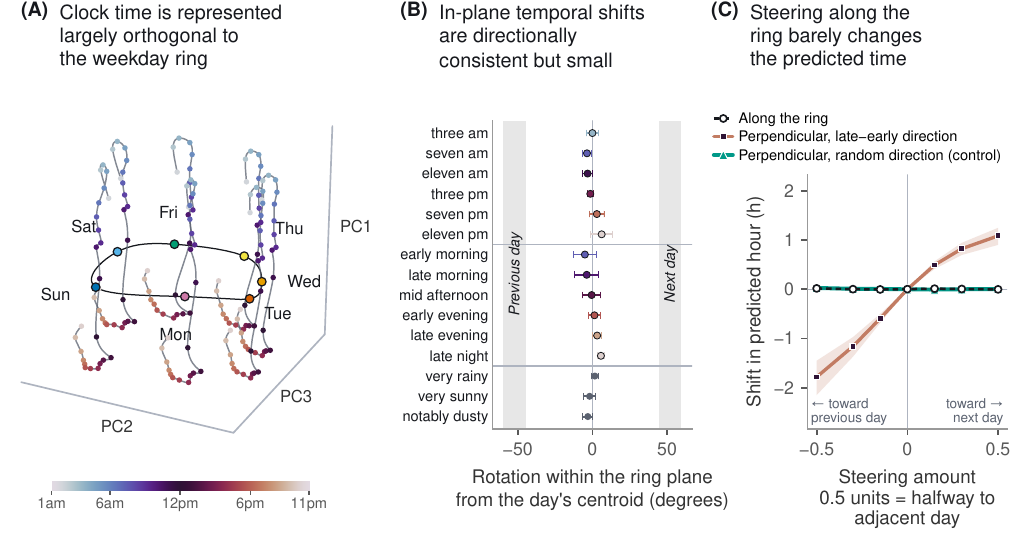}
\end{center}
\caption{\textbf{Time-of-day variation along the weekday ring is small and not causal} 
(A) Clock-time variation lies primarily along PC1, while weekday identity is captured mainly by PC2 and PC3. We show the top three principal components of weekday-token activations for prompts of the form ``It was \{time\} on \{day\}.'' 
(B) Time-of-day modifiers produce small (but systematic) angular shifts within the weekday plane, comparable in magnitude to those induced by non-temporal controls. Shifts are measured relative to each weekday centroid in the PC2--PC3 plane. Points show the mean across seven weekdays $\pm$ 1 s.e.m.
(C) Steering along the ring has almost no effect on the predicted hour, whereas steering along the off-ring component of the observed ``late--early'' displacement shifts predictions by one to two hours. The horizontal axis specifies the fraction of a weekday interval traversed along the ring. At each steering amount and layer, the off-ring late--early intervention (brown) and random orthogonal control (green) are matched in $\ell_2$ norm to the along-ring intervention (black). Curves show mean changes in predicted hour $\pm$ 95\% confidence intervals.}
\label{fig:timeofday}
\end{figure}

\subsection{Perturbation sensitivity has seven weekday-centred minima}
\label{sec:sensitivity}

Past work has shown that in-distribution activations can lie in regions where small perturbations have little downstream effect, separated by boundaries with much higher sensitivity to perturbations \citep{janiak2024stable}. This provides a way to probe the local structure of the weekday ring. If the fitted ring connects seven discrete weekday states, we might expect lower sensitivity near the weekdays and higher sensitivity in the interpolated regions between them. We test this using both finite steering and the local Jacobian and find that the weekday ring exhibits this predicted pattern.

First, we steer at constant speed around the fitted ring (similar to \citet{wurgaft2026manifold}) at layer-2 and measure the resulting behaviour and late-layer residuals. Equal movements along the ring have markedly different downstream effects. Both behaviour-space distance and late-layer residual distance change slowly around the weekdays and rapidly in the intervening arcs, producing a step-like progression around the ring (Figure~\ref{fig:sensitivity}A). We expect early and middle layers to represent the day of the week concept and late layers to represent the output prediction of the model. Accordingly, steering along the fitted ring closer to the readout (at layer-28) produces a smoother change in the behavioural output. We also measure how the step-like response from layer-2 steering develops across layers. The speed of the residual-stream response is lower near weekdays and higher in the intervening arcs, with the sevenfold contrast becoming more pronounced at progressively later layers (layers 3-31, Figure~\ref{fig:sensitivity}B). Thus, the nonuniform sensitivity develops within the network rather than arising solely at the final output readout.

Next, we measure local sensitivity directly using the Jacobian. Specifically, we compute the Jacobian of the layer-28 residual with respect to the layer-2 activation at states generated by steering across a polar grid spanning the weekday ring and its surrounding region. We use the Frobenius norm of this Jacobian as a direction-independent measure of perturbation sensitivity. Strikingly, the resulting sensitivity map has seven clear minima, each centred on a weekday representation. Moreover, when evaluated beyond the fitted ring, these low sensitivity regions extend radially from each weekday, forming seven spokes separated by regions of higher sensitivity (Figure~\ref{fig:sensitivity}C). The spoke structure therefore does not arise only from choosing a tangential path around the fitted ring (also see Appendix~\ref{app:models} where we explicity measure this).

To assess the robustness of the sevenfold sensitivity structure, we examine whether it depends on perturbation direction or experimental setup. In Llama-3.1-8B and Mistral-7B-v0.1, the pattern appears for tangential, radial, and orthogonal perturbations, whereas in GPT-2 XL it is most pronounced in the tangential perturbations (Appendix~\ref{app:models}). The pattern is also robust across prompt templates, including those used by \citet{wurgaft2026manifold}, and across readout positions, although its magnitude varies (Appendix~\ref{app:arithmetic}). 

One possible confound is that the discrete sensitivity pattern is driven by the existence of discrete weekday tokens. In fact we observe that the logits corresponding to the names of weekdays (as well as common abbreviations) vary significantly despite their probability being extremely small (Appendix~\ref{app:confound}). To examine this possibility we perform two controls (Appendix~\ref{app:confound}): First we set all weekday-related logits to -infinity and re-compute the logprobs. This indeed reduces the overall effect size, but a periodic oscillation aligned with the weekday spokes remains. We observe a similar lower-magnitude oscillation if we only measure the logit movement of a random set of tokens. Second, we test whether the most strongly-periodic tokens might be due to interference from the weekday tokens. To test this we measure the cosine similarity of these token unembeddings to the weekday unembedding vectors, and find them to be no more similar than random tokens.

Together, the finite steering and Jacobian measurements show that equal perturbations are not equally consequential around the fitted ring. Sensitivity is lowest around the seven weekday spokes and rises between them, a pattern consistent with seven discrete states rather than a single continuous coordinate.

\subsection{Time of day and uncertainty do not span the arcs between weekdays}
\label{sec:intermediate}

If the interpolated regions of the weekday ring correspond to meaningful intermediate values, there should be inputs whose representations occupy them. We test two plausible candidates: time within a day and uncertainty between candidate weekdays.

\subsubsection{Time-of-day variation along the weekday ring is small and not causal}
\label{sec:timeofday}

\begin{figure}[!t]
\begin{center}
\includegraphics[width=0.9\linewidth]{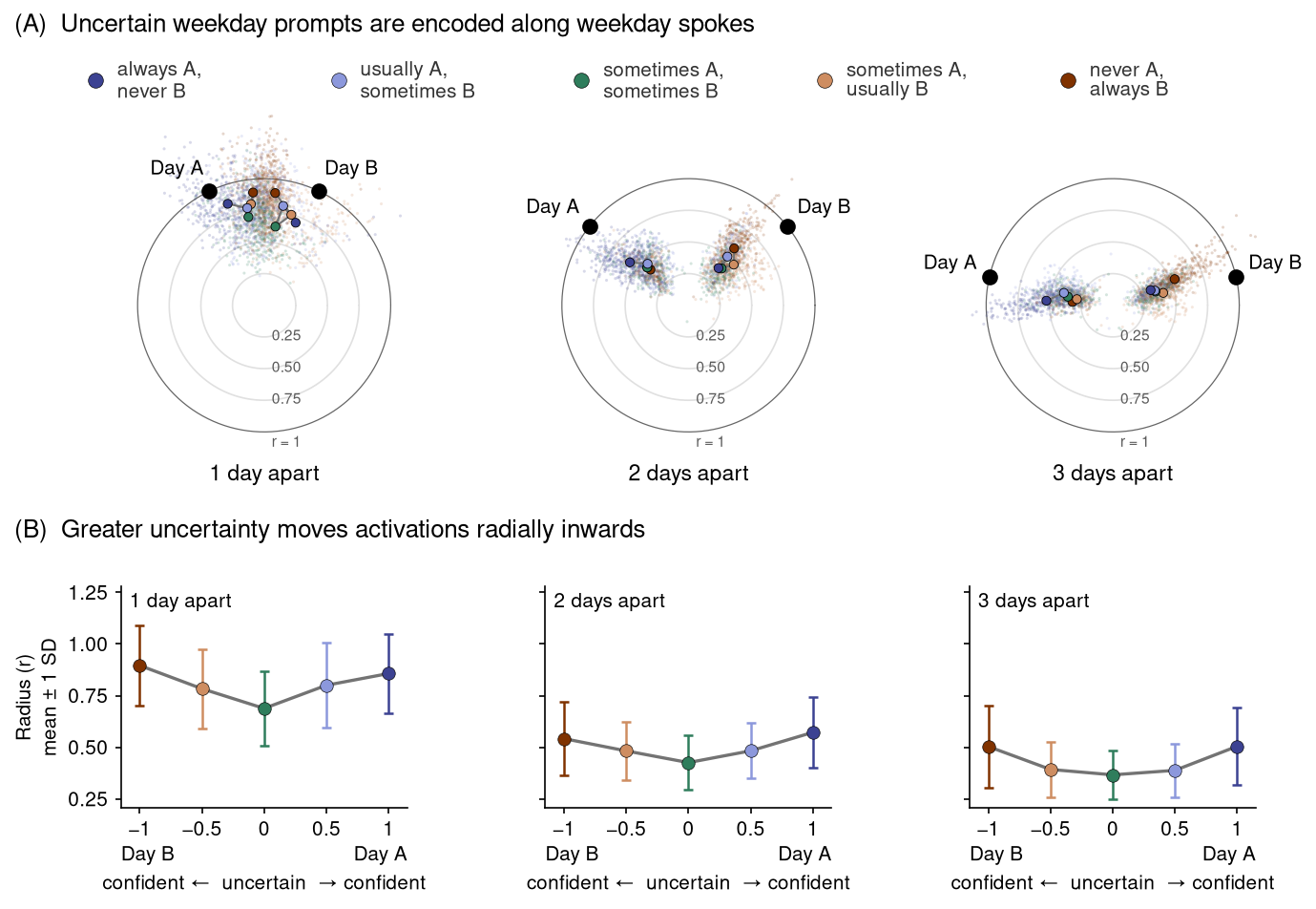}
\end{center}
\caption{\textbf{Uncertainty varies primarily along weekday spokes rather than between them.}
We vary the relative confidence assigned to two weekdays using frequency-adverb prompts, from ``Always on \{A\}, never on \{B\}'' through ``Sometimes on \{A\}, sometimes on \{B\}'' to ``Never on \{A\}, always on \{B\}''. Colours indicate the five confidence levels.
(A) Activations remain concentrated near the two candidate weekday spokes rather than filling the intervening arc. We show weekday pairs separated by one, two and three days. Small points represent individual activations; large points show the mean for each confidence level, computed separately for the two day-aligned groups. Concentric circles indicate normalised radius, with $r=1$ corresponding to the fitted ring.
(B) Activations move inwards as the stated confidence becomes more balanced, then outwards as the prompt increasingly favours the other weekday. The horizontal axis runs from full confidence in day B ($-1$), through equal confidence ($0$), to full confidence in day A ($+1$). Points show the mean normalised radius $\pm$ 1 SD.}
\label{fig:uncertainty}
\end{figure}

\citet{engels2025notall} report that time-of-day modifiers shift weekday representations towards neighbouring days, suggesting that the angular coordinate of the ring may encode time within a day. We test this directly using explicit clock times.

We prompt with ``It was \{time\} on \{day\}.'' for 22 clock hours and seven days (Appendix~\ref{app:timeofday}) and read the activations at the weekday token. PCA approximately factorises variation in clock time from variation in weekday identity:  PC1 mostly encodes clock time (99\% of its variance is between-hour), while the weekday ring lies in the PC2/PC3 plane (96\% of that plane's variance is between-weekday; Figure~\ref{fig:timeofday}A). Time-of-day variation is therefore nearly orthogonal to the weekday plane: 99.8\% of the variation among hour centroids lies outside it, and the leading clock-time axis makes an 88\textdegree{} angle with it.

However, a small component falls within the weekday plane and shifts representations (Figure~\ref{fig:timeofday}B). We quantify these shifts by measuring the angular displacement from each weekday centroid. We find that the shifts are systematic - times early in they day bias representations towards the preceding day and later times bias towards the next. Yet, these shifts are relatively small. For explicit clock times, the shifts are at most $6.30^\circ$, which is much smaller than the $44.33^\circ$--$59.81^\circ$ separating adjacent weekday centroids. We also find similarly small shifts for other temporal modifiers, such as ``early evening'' and ``late evening'' (similar to \cite{engels2025notall}). Furthermore, non-temporal controls such as ``very rainy'' produce shifts of similar magnitude, leaving unclear whether these small in-plane displacements carry time-of-day information.

To test whether this in-plane movement carries time-of-day information causally relevant to the model, we perform a steering experiment. We use prompts that require the model to predict an hour, such as ``The train leaves on Wednesday at around'', and compute the predicted hour from the next-token distribution at the end of the prompt. We steer the weekday-token activation along the fitted ring towards the next or previous day, comparing its effect with steering along the off-plane component of the observed ``late--early'' displacement, and along a random direction orthogonal to the ring (Appendix~\ref{app:timeofday}). To ensure a fair comparison, at each layer and steering amount, we rescale both off-plane interventions to match the $\ell_2$ norm of the displacement along the ring. 

Steering along the ring up to halfway towards a neighbouring day has almost no effect on the predicted hour ($ <0.02$ h, Figure~\ref{fig:timeofday}C). Similarly, a random orthogonal intervention of matched magnitude also has little effect ($ <0.03$ h). By contrast, steering by the same magnitude along the off-plane late--early direction shifts the predicted hour by $-1.8$h (towards ``early'') and $+1.1$h (towards ``late''), consistently across all 12 carrier prompts. Thus, although time-of-day modifiers induce a small displacement around the weekday ring, this in-plane component does not appear to carry time-of-day information used by the model.

\subsubsection{Uncertainty varies primarily along weekday spokes}
\label{sec:uncertainty}

\begin{figure}[!t]
\begin{center}
\includegraphics[width=\linewidth]{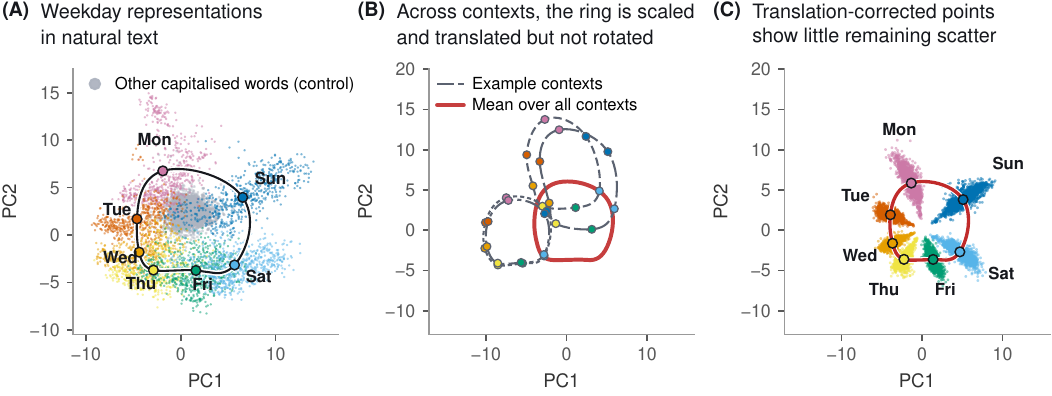}
\end{center}
\caption{\textbf{Natural context primarily translates weekday representations rather than moving them around the ring.}
(A) Representations of the same weekday are widely dispersed across natural contexts. We project final-token activations from 16-token FineWeb excerpts into the weekday plane, coloured by weekday. Grey points show excerpts ending in other capitalised words; large markers denote weekday centroids, and the curve shows the fitted ring.
(B) Different contexts primarily translate (and scale) the seven weekday representations while largely preserving their relative arrangement and orientation. For each excerpt, we construct seven variants differing only in the final weekday. Dashed curves show the resulting weekday configurations for four example contexts (footnote~\ref{fn:contexts}); the red curve shows the mean across contexts.
(C) Mean subtraction reveals that much of the apparent spread in (A) reflects context-dependent translation rather than movement into the arcs between weekdays. After subtracting the mean activation of the seven matched variants for each context, the representations form seven distinct weekday-aligned clusters. No rotational alignment is applied, indicating that the relative orientation of the weekday representations is largely preserved across contexts.}
\label{fig:occupancy}
\end{figure}

A second possibility is that positions between weekdays encode uncertainty between candidate days. The smooth redistribution of output probability between neighbouring weekdays reported by \citet{wurgaft2026manifold} is consistent with this interpretation. Prior work also shows that models can represent probability distributions over discrete latent states geometrically \citep{shai2024belief}. We therefore ask whether varying uncertainty between two candidate weekdays moves the representation through the interpolated region of the ring.

We construct prompts expressing graded uncertainty between pairs of weekdays separated by one, two, or three days, varying the stated probability or confidence assigned to each day. For example, one prompt reads ``Usually on Monday, sometimes on Tuesday, her flight leaves.'' We read the activation at the final period. Across several prompt families (Appendix~\ref{app:uncertainty}), we find that as the prompts express greater uncertainty, activations move inwards along one candidate weekday's spoke, rather than around the ring towards the other weekday (Figure~\ref{fig:uncertainty}). At the even split the mean radius is 21--27\% smaller than for the fully committed prompts (e.g. 0.56 $\to$ 0.43 for days two apart), while the angular offset from the nearest spoke is unchanged (0.19 vs 0.22 days). When the prompt favours the other weekday, 76--95\% of activations align with its spoke and move outwards as the stated confidence increases (Figure~\ref{fig:uncertainty}B). Thus, we infer that uncertainty varies along the weekday spokes rather than along the arcs between them. 

\subsection{Natural text does not systematically occupy interpolated regions}
\label{sec:naturaltext}

Finally, we analyse weekday representations in natural text to search more broadly for concepts that might be encoded in the interpolated regions of the ring. We sample 16-token excerpts from FineWeb, each ending in a weekday. We read the activation at the final token, where the circular geometry is the most pronounced (Appendix~\ref{app:ringtrace}).  We use a subset of these excerpts to identify the first two principal components capturing the weekday manifold's circular geometry and then project the remaining activations onto this two-dimensional subspace.

We observe that activations for the same weekday are widely dispersed across excerpts (Figure~\ref{fig:occupancy}A): the within-weekday scatter has an RMS radius of 0.65 ring radii, comparable to the 0.90-radius spacing between adjacent weekdays. This scatter could reflect a weekday-independent translation i.e. the preceding text shifting all seven weekday representations by a common vector, or a weekday-depenedent shift such as a rotation about the ring's centre. To distinguish between these, we mean-centre all activations: we replace the final weekday in each FineWeb excerpt with each of the other six (giving seven variants with identical preceding text) and subtract each excerpt's mean across its variants. This removes any shift shared by all weekdays, so the remaining scatter reflects weekday-dependent effects.

We find that preceding context primarily translates the seven weekday representations together, while preserving their relative arrangement and orientation (Figure~\ref{fig:occupancy}B).\footnote{\label{fn:contexts}The four example contexts in Figure~\ref{fig:occupancy}B, with the original day in braces, drawn as long dash, dash, short dash and dash-dot respectively: (1)~\emph{share the garden with your friends. The Daily Devotional is posted everyday, \{Monday\}}; (2)~\emph{the Product during business hours (between 9am and 5pm, \{Monday\}}; (3)~\emph{, president and chief executive of the American Petroleum Institute, said in a statement \{Thursday\}}; (4)~\emph{,'' Geng Shuang, a Chinese foreign ministry spokesperson, told reporters on \{Tuesday\}}.} After subtracting the mean activation of the seven variants for each context, the representations collapse into seven weekday-aligned clusters (Figure~\ref{fig:occupancy}C), the within-weekday scatter falling to 0.23 ring radii. Thus, we conclude that natural context primarily translates weekday representations rather than systematically rotating them into interpolated positions around the ring.

\section{Discussion}
\label{sec:discussion}

\subsection{Relation to prior evidence for continuity}
Two prior results have been interpreted as evidence that the weekday ring is continuous. \citet{wurgaft2026manifold} find smooth behavioural changes when steering along a fitted weekday manifold, while \citet{engels2025notall} find that temporal modifiers shift weekday representations towards neighbouring days. Our results qualify both interpretations.

We find that the smooth behaviour reported by \citet{wurgaft2026manifold} only occurs when interpolating at a late layer, close to the output. At such late layers, changes in the representation are mapped relatively directly into changes in the logits. By contrast, steering early allows differences between positions along the ring to be amplified or suppressed by subsequent computation. On the same prompts, steering the weekday token at layer 2 produces the step-like behaviour shown in Figure~\ref{fig:sensitivity} (Appendix~\ref{app:arithmetic}). This becomes more pronounced as the distance between the steering and readout layers increases. Thus, smooth behaviour under late-layer steering does not necessarily imply a continuous underlying representation.

\citet{engels2025notall} interpret shifts induced by temporal modifiers (e.g. ``early morning'') as evidence that the weekday ring may represent time continuously between neighbouring days. We reproduce these in-plane shifts, but find that (i) they are small relative to the separation between weekdays and similar in magnitude to shifts induced by non-temporal modifiers (e.g. ``notably dusty''), and (ii) that these in-plane shifts do not causally influence the model's time-of-day representations. Together, these results weaken the interpretation that the observed in-plane shifts reflect time-of-day being encoded continuously around the weekday ring.

\subsection{Scope and limitations}
Our results do not imply that the circular weekday geometry is spurious or meaningless. Indeed, recent work shows that cyclical feature arrangements can arise from correlations in the training data \citep{prieto2026data}. Instead, our claims concern the interpretation of this geometry as a continuous representation variable. Our results provide evidence for the view that the regions between the seven weekday representations do not correspond to meaningful intermediate representations.

To define the weekday manifold, we deliberately adopt the periodic-spline construction used by \citet{wurgaft2026manifold}, rather than introducing a new interpolation procedure. One could argue, that the higher sensitivity between weekday centroids is specific to this interpolation procedure. However, this possibility is itself relevant to interpreting fitted representation manifolds and central to the overarching goal of this study.  

The sevenfold sensitivity pattern we observe is consistent with discrete weekday states. Weekday-related logits contribute substantially to this pattern, although they do not fully explain it (Appendix~\ref{app:confound}). While prior work provides strong evidence that in-distribution activations lie in regions of low sensitivity, further work is needed to understand what causes activation plateaus and whether low sensitivity reliably identifies such regions. This limits how strongly we can interpret the sensitivity pattern as evidence for discrete states.

Finally, our work focused on the weekday ring; a single, particularly well-studied example of a proposed representation manifold. Because the weekday concept takes on a discrete set of values, its representations may be especially likely to exhibit signatures of discreteness despite the circular geometry. We leave extending these analyses to continuously varying concepts, such as colour, for future work.

\subsection{Concluding remarks}
Taken together, our results favour interpreting the weekday geometry as seven discrete spokes rather than a continuous ring. Although the weekday representations form a striking circular arrangement, our analyses of perturbation sensitivity and activation occupancy find little evidence that the arcs of the weekday ring encode meaningful intermediate states. Our findings call for caution when interpreting fitted manifolds, as their smooth geometry does not necessarily imply meaningful continuity in the underlying representation.


\ificlrfinal
\subsubsection*{Acknowledgements}
We are thankful to Josh Engels, Thomas Fel, Eric Michaud and Francisco Ferreira da Silva for helpful comments on the draft. We would also like to thank Philipp Kreer and other members of the Pivotal fellowship and LISA for useful discussions. This research was completed during the 2026 Q3 Pivotal Technical AI Safety Fellowship.
\fi

\ificlrfinal
\subsubsection*{Author Contributions}
DG and RF conducted the initial experiments that motivated the central hypothesis of the paper: DG on sensitivity measurements, RF on time of day shifts and ring occupancy. DG led the subsequent experimental work, including further sensitivity experiments, uncertainty experiments, extensions of time-of-day and natural-text analyses and the additional control analyses presented in the appendix. DG led the writing and revision of the manuscript. RF contributed to initial drafts of the manuscript. SH conceived and supervised the project and provided feedback on the manuscript.
\fi

\subsection*{AI use statement}

We used generative AI tools to implement methods (writing analysis and plotting code). We did not use them to develop hypotheses, design experiments, or interpret results; the remaining required-disclosure tasks are not applicable. We also used them for tasks with recommended disclosure: creating figures, drafting and editing the manuscript text, formatting references, and literature search. All AI-generated code was tested by the authors, figures were checked against the data, and AI-suggested references were verified against the original sources. We take responsibility for the final content of this work.

\subsection*{Ethics statement}

This work analyses open-weight language models on templated prompts and public web text (FineWeb), with no human subjects, private data, or model training. We see no direct harmful applications. As with other interpretability and steering research, the methods are dual-use. FineWeb text may contain biased or offensive content; we use it only as model input and reproduce only the four excerpts in footnote~\ref{fn:contexts}. Downstream uses of interpretability findings are hard to predict, and we welcome scrutiny of implications we have not foreseen.

\subsection*{Reproducibility statement}

All models are open-weight (Llama-3.1-8B, Mistral-7B-v0.1, GPT-2 XL) and are loaded through TransformerLens (Appendix~\ref{app:setup}). The prompt families, manifold construction, steering procedure, behavioural readout, sensitivity measurements, time-of-day analysis and natural-text filtering pipeline are fully specified in Appendices~\ref{app:prompts}--\ref{app:naturaltext}, including the numerical precision used for each sweep. Code to reproduce all experiments and figures is available at \url{https://github.com/dikshagup/spokes-not-a-ring.git}.

\bibliography{iclr2027_conference}
\bibliographystyle{iclr2027_conference}

\appendix

\section{Methods}
\label{app:methods}

\subsection{Setup}
\label{app:setup}

We study three open-weight base models: Llama-3.1-8B, Mistral-7B-v0.1 and GPT-2 XL. All seven weekday names are represented by single tokens in each model. All are loaded through TransformerLens \citep{nanda2022transformerlens}. Models are steered at layer 2 and read at $0.875$ of depth (layer 28 in Llama and Mistral, layer 42 in GPT-2 XL) unless otherwise specified.

\subsection{Prompts}
\label{app:prompts}

We use several prompt families to construct the weekday manifold and test its continuity. For the main sensitivity experiments, we use 20 day-neutral, third-person sentence templates, each instantiated with all seven weekdays, giving 140 prompts. For example, ``On Monday, she missed her train.'' We steer at the weekday token and measure downstream responses at the sentence-final period. We also use 49 weekday-arithmetic prompts of the form ``Q: What day is two days after Monday?\textbackslash nA:'', following \citet{wurgaft2026manifold}.

For the time-of-day experiments, we use prompts of the form ``It was \{time\} on \{day\}.'', varying the clock hour across all seven weekdays. We additionally test coarse time-of-day expressions and non-temporal weather and mood modifiers as controls, for example,``It was late night on Monday.'', ``It was notably dusty on Monday.'' Every modifier is exactly two tokens. To measure causal effects on hour predictions, we use twelve sentence stems ending immediately before an hour, such as ``The train leaves on Wednesday at around''.

For the uncertainty experiments, we construct six prompt families expressing different degrees of confidence between two weekdays. The main family uses frequency adverbs, for example ``Usually on Monday, sometimes on Tuesday, her flight leaves.'' The other families express confidence through alternative adverbs, verbal odds, percentages and either/or phrasing. We vary the weekday pair, naming order and sentence continuation within each family (Appendix~\ref{app:uncertainty_methods}).

Finally, the natural-text experiment uses excerpts sampled from FineWeb (Appendix~\ref{app:naturaltext}).

\subsection{The activation manifold}
\label{app:manifold}

The manifold $\Mh$ at site $h$ is the closed curve through the seven weekday centroids: a periodic cubic spline fit per coordinate with periodic boundary conditions, parameterised by an intrinsic coordinate $u \in [0, 1)$ with knots $u_d = d/7$, interpolating the centroids exactly ($\sigma(u_d) = c_d$). Centroids are per-day means over that day's prompts. Centroids and spline are formed in the full residual stream ($\dmodel$ dimensions). For every fit manifold, we confirmed that the minimum-length cyclic tour of the seven centroids under Euclidean distance recovers the true calendar order. Principal components are primarily used for visualisation.

\subsection{Activation steering}
\label{app:steering}
\begin{figure}[h]
\begin{center}
\includegraphics[width=\linewidth]{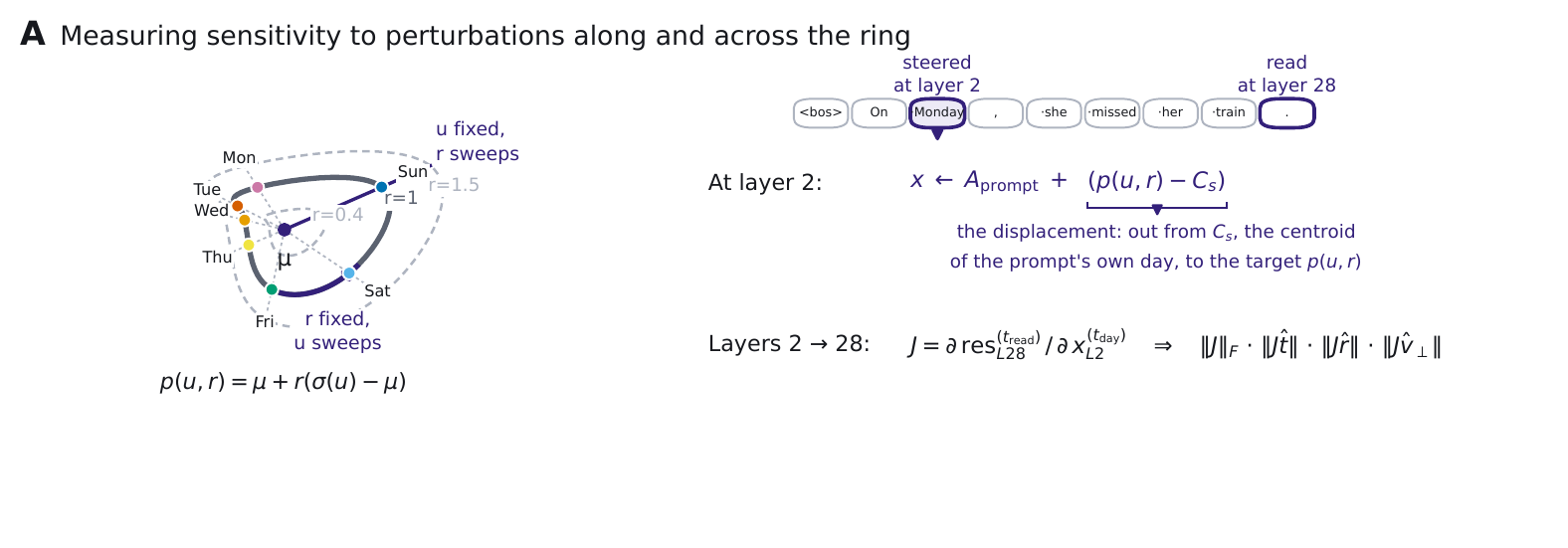}
\end{center}
\caption{Schematic of the steering construction and sensitivity analysis.}
\label{fig:schematic}
\end{figure}

We steer along the manifold with an activation-addition write \citep{turner2024steering} at a single token position. For a prompt $\rho$ whose stated day is $s$, steering along the manifold to intrinsic coordinate $u$ replaces the recorded activation $a_\rho$ with
\begin{equation}
  x = a_\rho + \sigma(u) - c_s,
  \label{eq:steer}
\end{equation}

where $c_s$ is the centroid of the prompt's own day and $\sigma(u)$ is the spline evaluated at $u$. Every term is a vector in $\mathbb{R}^{\dmodel}$. At the knots, this reduces exactly to a difference of day-conditional means, $c_t - c_s$, while intermediate $u$ gives a continuous sweep along the ring.

For steering along different parts of the ring, we write $\mu = \frac{1}{7}\sum_d c_d$ for its centre and replace the target $\sigma(u)$ with the polar point
\begin{equation}
  p(u, r) = \mu + r\,(\sigma(u) - \mu),
  \label{eq:polar}
\end{equation}
recovering Eq.~\ref{eq:steer} at $r = 1$, the ring centre at $r = 0$, and points outside the ring at $r > 1$. This is the target used for the Jacobian field and the radial sweep (Figure~\ref{fig:schematic}).

\subsection{Behavioural readout}
\label{app:readout}

Following \citet{wurgaft2026manifold}, the behaviour manifold $\My$ lives in the space of output distributions. The next-token distribution at the capture position is restricted to the seven weekday tokens and an `other' class holding the remaining mass, giving a point on the 8-simplex. The simplex is mapped into Hellinger space by $p \mapsto \sqrt{p}$, in which
\begin{equation}
  d_H(p, q) = \frac{1}{\sqrt{2}}\,\big\|\sqrt{p} - \sqrt{q}\big\|_2
\end{equation}
is an ordinary Euclidean distance. Path lengths are cumulative sums of per-step distances.

\subsection{Sensitivity measurements}
\label{app:sensitivity}

We measure how fast the readout moves as the steered activation moves in two independent ways.

\paragraph{Finite differences.}
While steering (Appendix~\ref{app:steering}), we record the residual step at depth $\ell$ divided by the input step,
\begin{equation}
  s_\ell(u) = \frac{\|\mathrm{res}_\ell(u+1) - \mathrm{res}_\ell(u)\|}{\|x(u+1) - x(u)\|},
\end{equation}
with $\mathrm{res}_\ell$ the full-dimensional residual at the readout position and depth $\ell$, and the denominator the step at the patched position. Each run patches a single layer and a single position. The numerator is a small difference of two large, nearly equal vectors, so it is sensitive to arithmetic precision. All reported sweeps use float16 (Llama, Mistral) or float32 (GPT-2 XL).

\paragraph{The Jacobian.}
We evaluate $J_\ell = \partial\,\mathrm{res}_\ell / \partial x$ at each steered state by forward-mode automatic differentiation, on a polar grid over the day disc: 56 samples around the ring and 17 radii out to $r_{\max} = 1.6$ in steps of $0.1$, over 70 prompts (10 templates; Figure~\ref{fig:schematic}). We compute the direction-free norm $\|J_\ell\|_F$, estimated by Hutchinson's estimator \citep{hutchinson1989stochastic}.

We also compute three directional gains recorded at every grid point: the tangential and radial gains $\|J_\ell \hat{t}\|$ and $\|J_\ell \hat{r}\|$, and a control gain $\|J_\ell \hat{v}_\perp\|$ along a random unit direction orthogonal to the six-dimensional centroid span, drawn isotropically, projected and renormalised.

\subsection{Time of day}
\label{app:timeofday}

We analyse the weekday-token activations for the clock-time and temporal-modifier prompts described in Appendix~\ref{app:prompts}. Since the time expression precedes the weekday, its information is available at the readout position. We fit PCA to the 210 clock-time prompts and subsequently project the non-temporal controls into the same space.

We characterise each principal component by the proportion of its variance attributable to clock time or weekday identity. PC1 explains 19.5\% of the total variance, of which 99\% is between-hour variation. PC2 and PC3 explain 13.5\% and 11.5\%, respectively, and predominantly capture weekday identity (91\% and 97\% between-weekday variation). We therefore define the weekday plane $P$ as the span of PC2 and PC3. We compute a centroid for each weekday by averaging its activations and fit a periodic cubic spline through the seven centroids in weekday order (Appendix~\ref{app:manifold}). For visualisation, we connect the clock-time activations for each weekday, ordered by hour, using a natural cubic spline.

To quantify time-of-day shifts within the weekday plane, we project each activation onto $P$ and measure its angular displacement $\Delta\theta$ from the corresponding weekday centroid, wrapped to $\pm180^\circ$. For each modifier, Figure~\ref{fig:timeofday}B reports the mean displacement across the seven weekdays, with error bars showing $\pm1$ standard error ($n=7$). For comparison, adjacent weekday centroids are separated by $44.5^\circ$--$59.6^\circ$ in this plane.

\paragraph{Causal steering.}
We use prompts of the form ``\{carrier\} on \{day\} at around'', such as ``The train leaves on Wednesday at around''. We select twelve of 32 candidate sentence stems based on their unsteered hour distributions, excluding those strongly concentrated on a single hour (concentration $>0.55$) to allow measurable changes under steering. Combined with all seven weekdays, these give 84 prompts. We estimate the predicted hour by aggregating probabilities across 625 written forms of clock times, such as ``7pm'', ``7 p.m.'' and ``19:00'', using continuations to disambiguate shared prefixes. We then compute the probability-weighted circular mean over the 24 hours.

We compare steering along three directions at the weekday token. At each layer $\ell=2,\ldots,28$, we fit a periodic spline through the seven full-dimensional weekday centroids obtained from neutral prompts. All seven centroids retain their calendar order at every layer. The first intervention moves the activation along this fitted ring towards the next or previous weekday. For the second, we compute the mean ``late--early'' activation difference for each weekday and remove its component along the ring's local tangent, obtaining an off-ring direction. The third uses a random direction orthogonal to the same tangent as a control.

We apply all three interventions at signed steering amounts $\alpha \in \{0,\pm0.15,\pm0.3,\pm0.5\}$. For along-ring steering, $\alpha$ specifies the fraction of a weekday interval traversed from the clean activation's nearest point on the spline, while preserving its displacement off the spline. At each layer, we scale the off-ring and random interventions to match the $\ell_2$ norm of the corresponding along-ring displacement. Intervention targets are defined relative to the clean activation at each layer, preventing the applied displacements from accumulating across layers. We report the change in predicted hour relative to the unsteered run, averaged over all 84 prompts, with 95\% confidence intervals from 2,000 bootstrap resamples over the twelve carriers.

\subsection{Uncertainty between two weekdays}
\label{app:uncertainty_methods}

\paragraph{Prompts.}
We construct prompts expressing graded uncertainty between two weekdays. Our main family uses frequency adverbs, for example ``Usually on Monday, sometimes on Tuesday, her flight leaves.'' We vary the relative confidence assigned to the two days across five levels: \emph{always/never}, \emph{usually/sometimes}, \emph{sometimes/sometimes}, \emph{sometimes/usually} and \emph{never/always}. We combine these with all 21 unordered weekday pairs (seven at each circular separation of one, two and three days), both naming orders and 32 third-person continuations describing scheduled events, yielding 6,720 prompts.

To test robustness to phrasing, we construct five additional families using certainty adverbs, alternative frequency adverbs, verbal odds, percentages and either/or expressions. We combine these with the same 21 weekday pairs, both naming orders and twelve continuations spanning first-, second- and third-person sentences. Within each family, all confidence modifiers have the same token length, keeping the weekday-token positions fixed across confidence levels. For all six families, we read the layer-28 residual-stream activation at the sentence-final period in Llama-3.1-8B.

\paragraph{Ring coordinates.}
We project each activation into the reference weekday subspace and express its position using an angular coordinate and a normalised radius. Let $\sigma(u)$ denote the fitted ring and $\mu$ its centre. For a projected activation $z$, we choose the coordinate $u$ whose spoke $\sigma(u)-\mu$ is most closely aligned with $z-\mu$. We then measure how far the activation lies along that spoke, normalised by the distance from the centre to the ring:
\begin{equation}
    r = \frac{(z-\mu)\cdot\hat{s}(u)}
             {\|\sigma(u)-\mu\|},
    \qquad
    \hat{s}(u) =
    \frac{\sigma(u)-\mu}{\|\sigma(u)-\mu\|}.
\end{equation}
Thus, $r=0$ corresponds to the centre and $r=1$ to the fitted ring. This approach follows the spoke directions defined by the polar steering map (Eq.~\ref{eq:polar}), rather than projecting onto the nearest point of the spline, which can give ambiguous angular positions near the centre.

To compare activations across weekday pairs, we express their angular positions relative to the two days named in each prompt. Let $u_A$ and $u_B$ denote the corresponding weekday coordinates, with day A preceding day B under our weekday-ordering convention. Using the shortest signed circular differences, we define
\begin{equation}
    w = \frac{u-u_A}{u_B-u_A}.
\end{equation}
Here, $w=0$ corresponds to day A's spoke, $w=1$ to day B's spoke and $w=0.5$ to the midpoint of the intervening arc.

We also assign each prompt a signed preference score, ranging from $+1$ when day A is most strongly favoured, through $0$ for equal preference, to $-1$ when day B is most strongly favoured. For the main prompt family, we divide the confidence-level rank by two. For the remaining families, we use $2p-1$, assigning the sign according to which weekday receives the stronger modifier, where $p$ is the stated probability of the favoured day.

\subsection{Natural text in the weekday subspace}
\label{app:naturaltext}

We analyse naturally occurring weekday representations to determine whether apparent occupancy of the arcs between weekdays reflects movement around the ring or context-dependent translation.

We sample text from FineWeb sample-10BT \citep{penedo2024fineweb}, selecting excerpts that end in a weekday. We also collect excerpts ending in other capitalised words as controls. After removing overlapping and near-duplicate samples, we retain $49{,}678$ weekday excerpts from $574{,}217$ documents. We present the final 16 tokens of each excerpt to the model, preceded by a beginning-of-sequence token, and record the layer-28 activation at the final token. We use a document-disjoint subset of $16{,}879$ weekday excerpts to fit the weekday subspace and reserve the remaining $32{,}799$ for analysis.

\paragraph{Matched-context variants.}
To distinguish movement around the ring from context-dependent translation, we construct seven variants of each weekday-final excerpts by replacing its final token with each weekday while keeping the preceding text unchanged. We record the layer-28 activation for every variant. Each matched set therefore contains seven weekday representations elicited by the same preceding text. We subtract their mean to remove any common translation, without rotating the representations. Figure~\ref{fig:occupancy}B--C uses the $32{,}799$ held-out clean weekday contexts; the four examples in panel B are selected for their large translations within four predefined context categories.

We quantify how the seven-day arrangement changes across contexts by comparing each matched set with a reference arrangement, defined as the mean-centred weekday configuration from the fitting subset.

\clearpage
\section{Reproducing sensitivity results with other models}
\label{app:models}

\begin{figure}[!t]
\begin{center}
\includegraphics[width=0.85\linewidth]{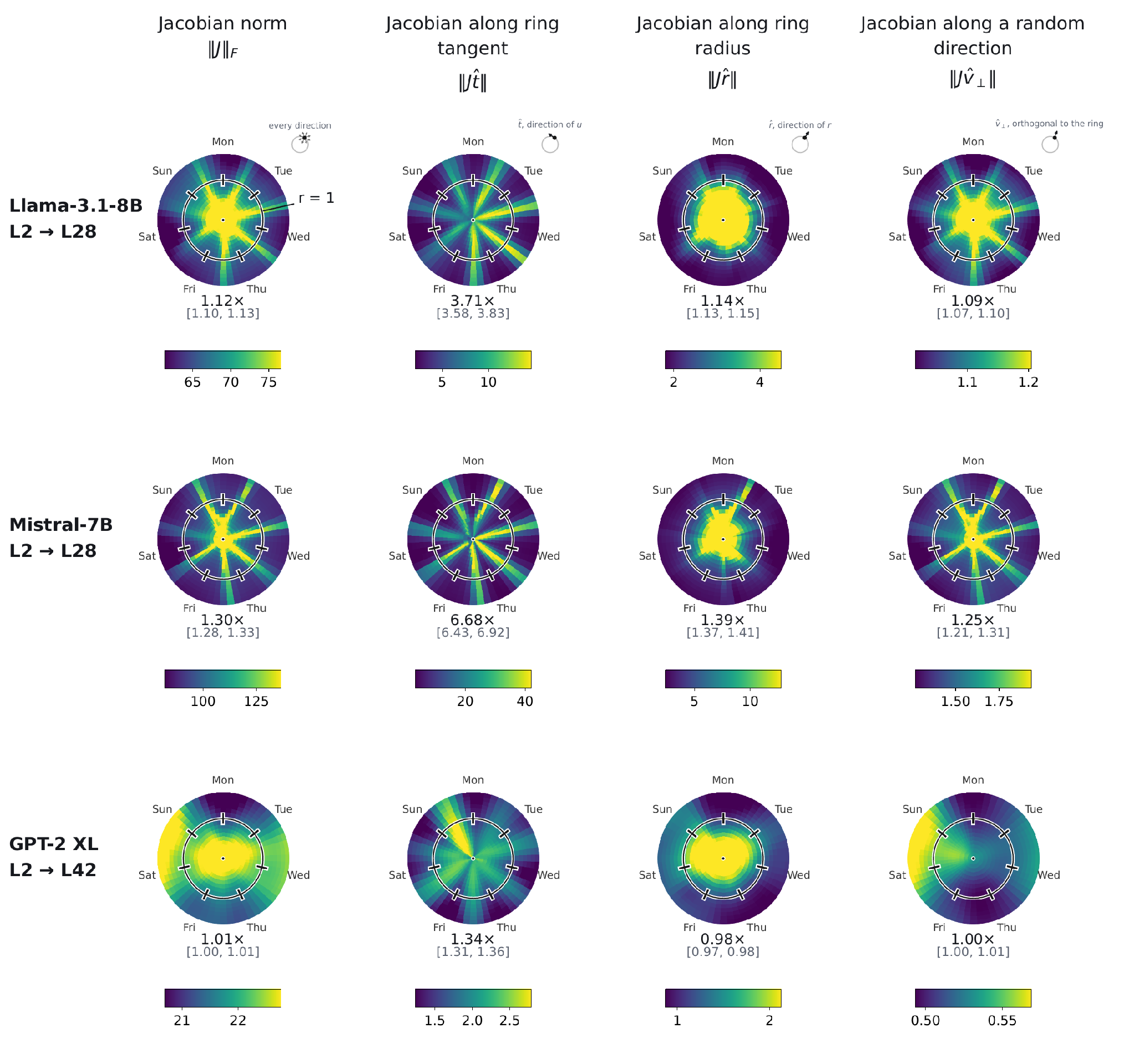}
\end{center}
\caption{The Jacobian of the readout on the weekday disc $p(u, r) = \mu + r(\sigma(u) - \mu)$, in three models. The number under each disc is how much more sensitive the readout is between the days than at them, on the ring itself ($r = 1$), with a 95\% bootstrap interval over background prompts.}
\label{fig:models}
\end{figure}

We measured Jacobians (Appendix~\ref{app:sensitivity}) in Llama-3.1-8B, Mistral-7B-v0.1 and GPT-2 XL on the day-neutral, third-person sentence templates (Appendix~\ref{app:prompts}). The effect is most pronounced along the tangent across models: $3.71\times$ in Llama, $6.68\times$ in Mistral, $1.34\times$ in GPT-2 XL. In Llama and Mistral, the same seven-fold spokes structure persists in the norm ($1.12\times$, $1.30\times$), the radius ($1.14\times$, $1.39\times$) and even along a random direction orthogonal to the ring span ($1.09\times$, $1.25\times$). However, in GPT-2 XL the other three sit at unity ($1.01\times$, $0.98\times$, $1.00\times$).

\clearpage
\section{Non-uniform sensitivity with weekday arithmetic}
\label{app:arithmetic}

\begin{figure}[!t]
\begin{center}
\includegraphics[width=0.8\linewidth]{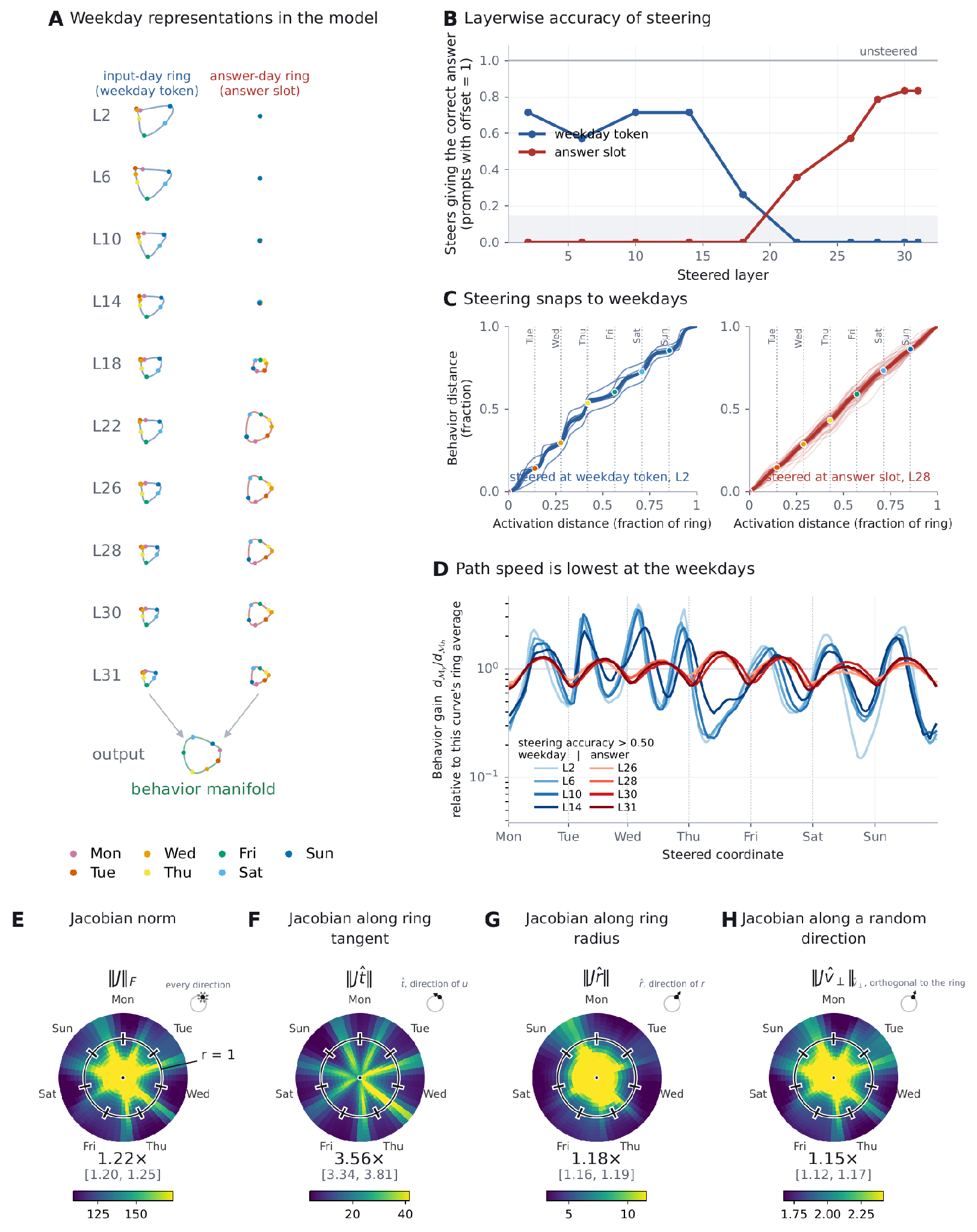}
\end{center}
\caption{Perturbation sensitivity of Llama-3.1-8B on arithmetic prompts from  \citealp{wurgaft2026manifold}. (A)~Day centroids at every layer for the weekday token (blue, index 9) and the answer slot (red, index 12), projected to three dimensions under one shared camera, ending in the behaviour manifold $\My$ (Appendix~\ref{app:readout}). (B)~Fraction of steers producing the correct answer, by steered layer, for each site; the shaded band is chance. (C)~On-manifold distance in activation space against on-manifold distance in behaviour space, after \citet{wurgaft2026manifold}, for the weekday token at L2 (left) and the answer slot at L28 (right). Dotted lines mark the day knots. (D)~Behaviour gain $d\My/d\Mh$ along the ring, relative to each curve's own ring average; only layers clearing $0.50$ steering accuracy in (B) are drawn. (E--H)~The Jacobian of the layer-28 readout with respect to the steered day token, measured as in Appendix~\ref{app:sensitivity} on the disc of Eq.~\ref{eq:polar}: Frobenius norm, then gain along the ring tangent, the radius, and a random direction orthogonal to the ring's span.}
\label{fig:arithmetic}
\end{figure}

The results in Section~\ref{sec:sensitivity} use day-neutral, third-person sentence templates. Here we repeat the measurements on interrogative prompts that require arithemetic: ``Q: What day is two days after Monday?\textbackslash nA:'' from \citet{wurgaft2026manifold}. The prompt has two candidate sites, at the weekday token and the answer token (Figure~\ref{fig:arithmetic}A): day recovery at the weekday token is $1.000$ from layer 2, while the answer slot is at $0.122$ at layer 2, below the $1/7$ chance level. Answer-slot accuracy does not exceed $0.80$ until layer 18. This is mirrored in the steering accuracy (Figure~\ref{fig:arithmetic}B): steering along the manifold at the weekday token succeeds at roughly $0.7$ through layer 14 and falls to zero by layer 22, while steering along the answer manifold stays at zero until layer 18 and reaches $0.83$ by layer 31.

The step-like behavioural response and the spokes pattern in Jacobian norm (Figure~\ref{fig:sensitivity}) reproduce here. Sweeping the ring at the weekday token, behaviour distance advances in steps, flat at each day and steep across the arcs, while the same sweep at the answer slot is smooth (Figure~\ref{fig:arithmetic}C). This is consistent with the finding that plateaus sharpen with the number of layers between perturbation and readout \citep{shinkle2025plateaus}. Similarly, the corresponding behaviour gain varies with minima at the seven weekday centroids, at the weekday token, against a much flatter profile at the answer slot (Figure~\ref{fig:arithmetic}D). The Jacobian profiles reproduce the spokes: $1.22\times$ for the Frobenius norm, $3.56\times$ along the ring tangent, $1.18\times$ radially and $1.15\times$ along a random direction orthogonal to the ring span (Figure~\ref{fig:arithmetic}E--H). The layer dependence we observe is worth noting alongside the isometry reported by \citet{wurgaft2026manifold}, which is measured between centroids at the last token of layer 28 at the answer slot, where the mapping is closest to uniform.

\clearpage
\section{Weekday-related logits do not explain the sevenfold sensitivity pattern}
\label{app:confound}

\begin{figure}[!t]
\begin{center}
\includegraphics[width=0.9\linewidth]{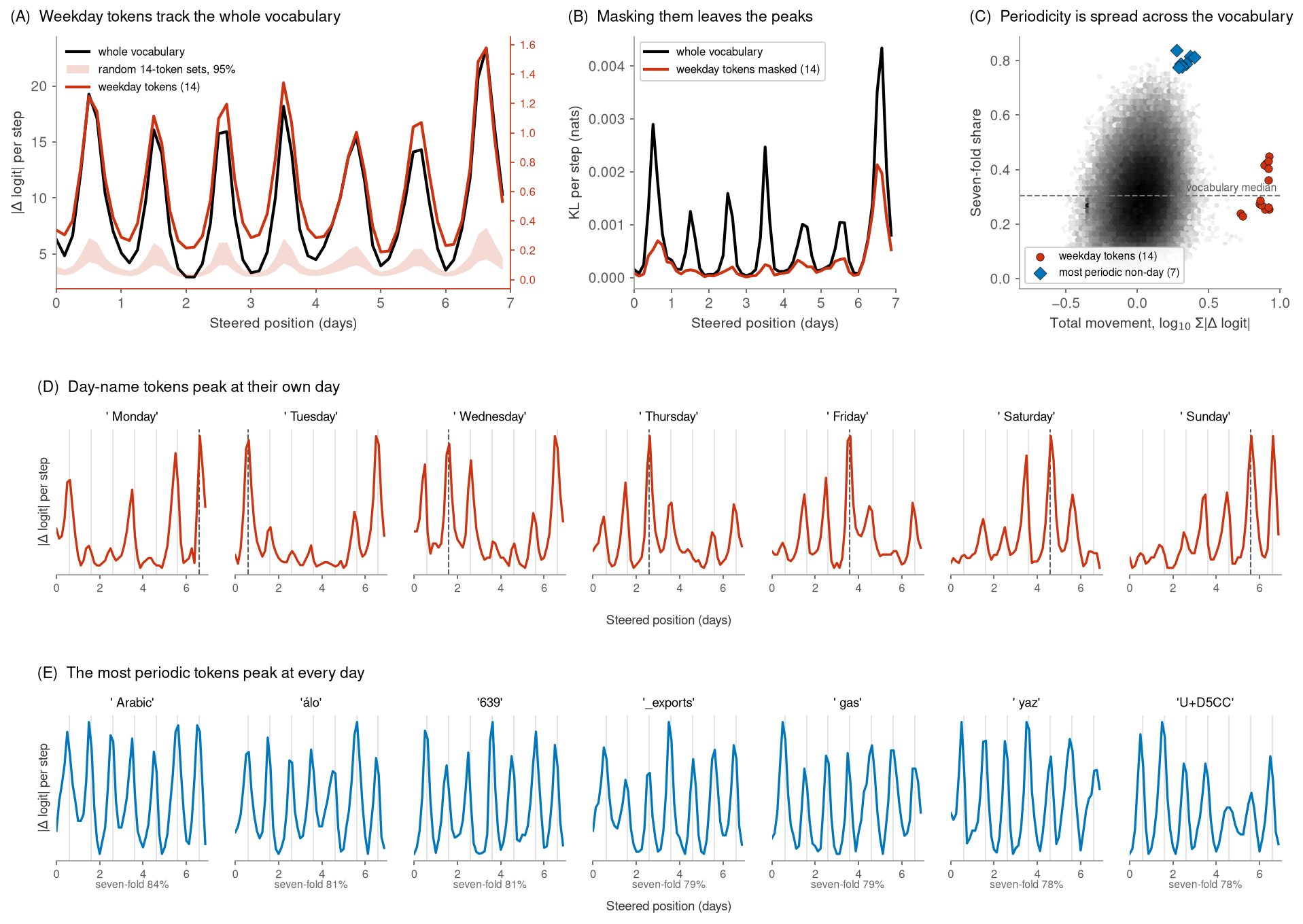}
\end{center}
\caption{\textbf{The sevenfold sensitivity pattern is not driven by weekday-related logits.} Llama-3.1-8B, mention family, steering around the ring at layer 2 and reading the output logits. (A)~Mean absolute logit change per steering step for the whole vocabulary (black, left axis) and for the 14 weekday tokens (red, right axis); the shaded band is the 95\% range over random 14-token subsets. Both track the same sevenfold profile. (B)~KL divergence per step of the output distribution with the weekday tokens included (black) and with their logits set to $-\infty$ before the softmax (red). Masking reduces the overall signal but leaves the seven peaks. (C)~For every vocabulary token, the share of its logit-change spectrum at the sevenfold frequency against its total movement. Weekday tokens (red) move most but have near-median periodicity; the most periodic tokens (blue) are non-weekday tokens. (D)~Per-step logit change for each day-name token: each peaks at its own day (dashed line) rather than showing sevenfold periodicity. (E)~The seven most periodic tokens in the vocabulary, none of which is weekday-related, peak at every day.}
\label{fig:confound}
\end{figure}

Section~\ref{sec:sensitivity} raises the possibility that the sevenfold sensitivity pattern is driven by weekday-related logits, since steering around the ring strongly changes these logits. Figure~\ref{fig:confound} tests this in two ways. First, we set the logits of the 14 weekday tokens (each day with and without a leading space) to $-\infty$ before computing the output distribution. This substantially reduces the overall signal, but the sevenfold modulation remains (Figure~\ref{fig:confound}B). Second, we examine individual logits. Weekday logits show large day-specific changes rather than sevenfold periodicity: each day-name token peaks near its own day (Figure~\ref{fig:confound}D). By contrast, sevenfold periodicity is spread across the vocabulary, appears in randomly sampled token subsets (Figure~\ref{fig:confound}A) and is strongest in non-weekday tokens (Figure~\ref{fig:confound}C, E). We confirm that these periodic tokens are no more similar to the weekday unembedding vectors than random vocabulary tokens. Thus, weekday-related outputs amplify the sensitivity pattern but do not explain its sevenfold structure.

\clearpage
\section{Uncertainty between weekdays across prompt families}
\label{app:uncertainty}

\begin{figure}[H]
\begin{center}
\includegraphics[width=\linewidth]{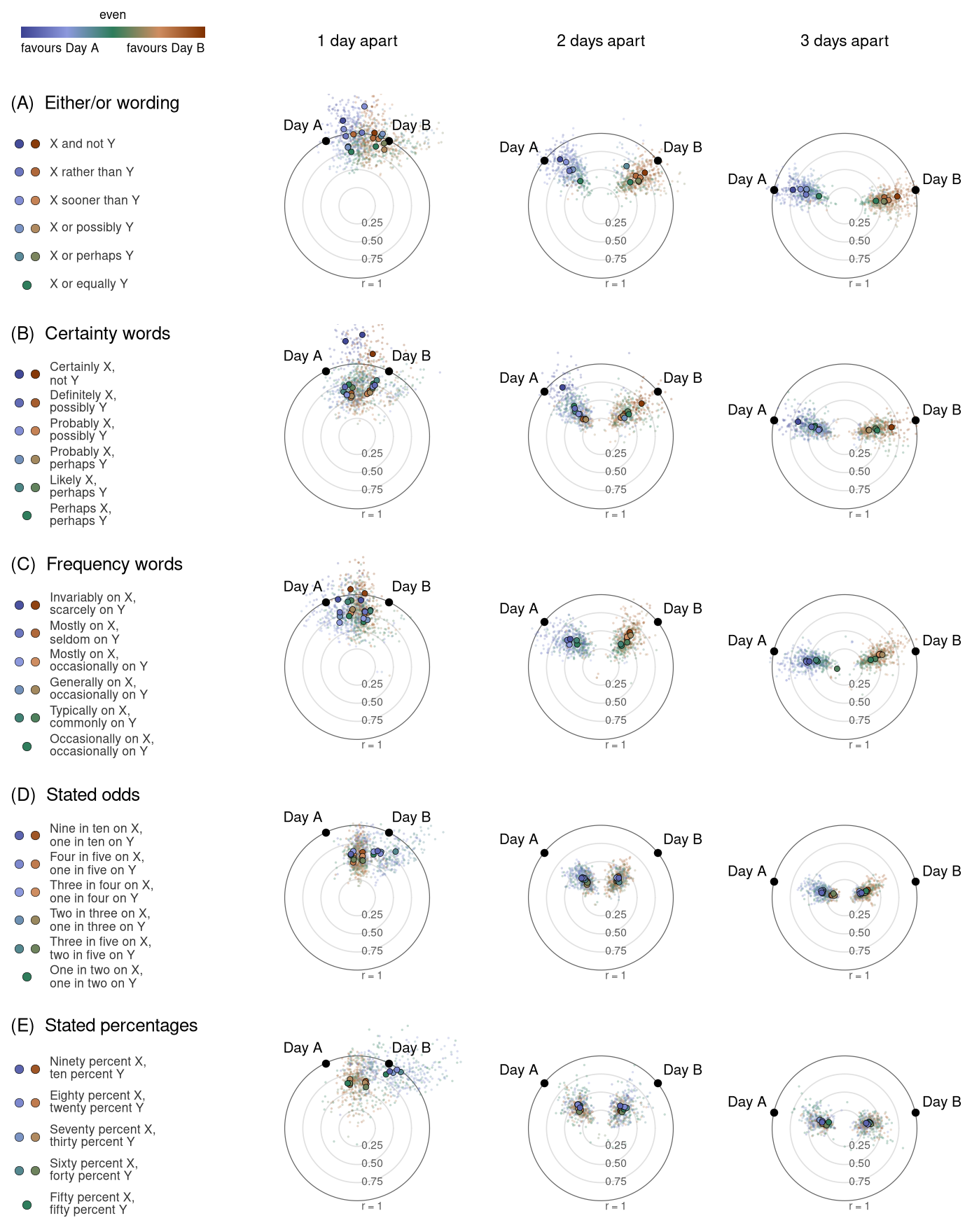}
\end{center}
\caption{\textbf{Uncertain weekday prompts are encoded along weekday spokes across five prompt families.} Each row is a prompt family expressing graded preference between two candidate weekdays, with the wording of each level listed on the left; columns show pairs one, two, or three days apart. Final-token activations are projected into the weekday plane and coloured from favouring day A (blue) through even (green) to favouring day B (brown); large markers show the mean at each level after separating the two modes of the cloud. (A)~Either/or wording. (B)~Certainty words. (C)~Frequency words. (D)~Stated odds. (E)~Stated percentages.}
\label{fig:uncertainty_families}
\end{figure}

Section~\ref{sec:uncertainty} reports results for one prompt family. Here we repeat the analysis for five families that express graded uncertainty between two candidate weekdays in different ways: either/or wording (``X and not Y'' through ``X or equally Y''), certainty words (``Certainly X, not Y'' through ``Perhaps X, perhaps Y''), frequency words (``Invariably on X, scarcely on Y'' through ``Occasionally on X, occasionally on Y''), stated odds (``Nine in ten on X, one in ten on Y'' through ``One in two on X, one in two on Y''), and stated percentages (``Ninety percent X, ten percent Y'' through ``Fifty percent X, fifty percent Y''). Each family is run in both orders of the two days, for pairs one, two, or three days apart (Figure~\ref{fig:uncertainty_families}).

The pattern of Figure~\ref{fig:uncertainty} holds in every family. For pairs two or three days apart, activations form two clusters aligned with the two candidate weekday directions rather than a continuum through the arc between them, and the clusters move inward as the stated uncertainty increases. The families differ mainly in radius: numerical wordings (stated odds and percentages) sit closer to the centre of the ring than the lexical wordings, but remain aligned with the weekday spokes. For adjacent days the two spokes are too close for the clusters to separate cleanly.

\clearpage
\section{The weekday ring across read positions and layers with natural text}
\label{app:ringtrace}

\begin{figure}[H]
\begin{center}
\includegraphics[width=0.9\linewidth]{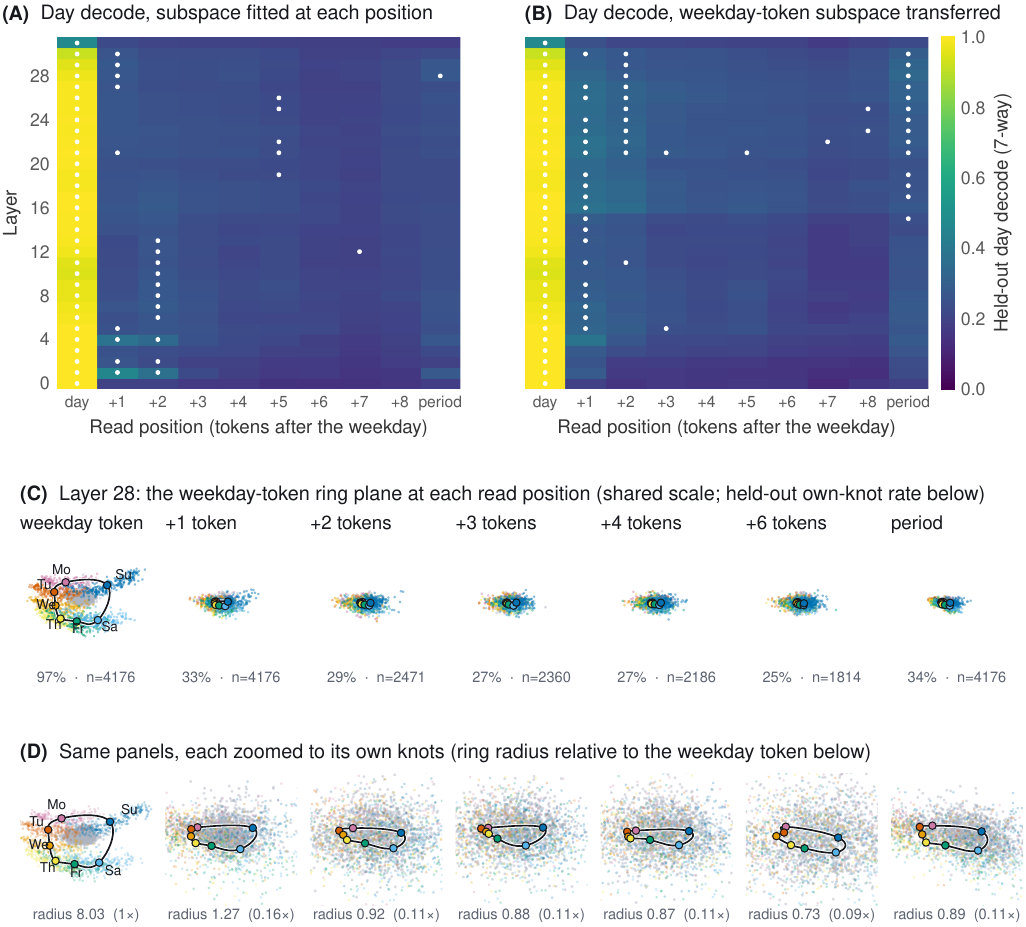}
\end{center}
\caption{\textbf{The weekday ring is sharpest at the weekday token and decays at later read positions.} Llama-3.1-8B, FineWeb excerpts. (A)~Held-out seven-way day decoding accuracy at every layer and read position, with the weekday subspace fitted separately at each position; dots mark cells above chance decoding. (B)~The same, with the subspace fitted at the weekday token and transferred to every other position. (C)~Layer-28 activations projected into the weekday-token ring plane at each read position, on a shared scale; the held-out rate of projecting closest to the activation's own weekday centroid is given below each panel. (D)~The same panels, each zoomed to its own ring, with the ring radius relative to the weekday token given below.}
\label{fig:ringtrace}
\end{figure}

Figure~\ref{fig:ringtrace} traces the weekday ring across read positions and layers for the excerpts from FineWeb. Day identity is decodable at the weekday token from the earliest layers, with near-perfect held-out accuracy, whether the weekday subspace is fitted at that position (Figure~\ref{fig:ringtrace}A) or transferred from it (Figure~\ref{fig:ringtrace}B). At positions after the weekday, decodability drops sharply and remains near chance, rising only at a few sporadic layers. We project and visualise activations from layer-28 activations into the weekday-token ring plane. At later positions the seven weekday clusters collapse towards the centre, with the ring radius shrinking to roughly a tenth of its value at the weekday token (Figure~\ref{fig:ringtrace}C--D). This motivates our analysis at the weekday token in the main text.

\end{document}

%% file: math_commands.tex
\usepackage{amsmath,amsfonts,bm}

\def\eqref#1{equation~\ref{#1}}

\def\1{\bm{1}}

\DeclareMathAlphabet{\mathsfit}{\encodingdefault}{\sfdefault}{m}{sl}
\SetMathAlphabet{\mathsfit}{bold}{\encodingdefault}{\sfdefault}{bx}{n}

